\documentclass[letterpaper, 10 pt, conference]{ieeeconf}  
\usepackage[T1]{fontenc}
\usepackage{xcolor}
\usepackage{graphicx}
\usepackage{amsmath}
\usepackage{glossaries}
\usepackage{dblfloatfix}
\usepackage{url}
\usepackage{float}
\usepackage{amssymb}
\usepackage{yfonts}
\usepackage{booktabs}
\usepackage{siunitx}
\PassOptionsToPackage{hyphens}{url} 
\usepackage[unicode]{hyperref}
\hypersetup{hidelinks, breaklinks=true}
\usepackage{fancyhdr}

\let\labelindent\relax
\usepackage[inline]{enumitem}

\fancypagestyle{withfooter}{
  \fancyhf{}
  \fancyhead[C]{\textcolor{gray}{This paper has been accepted for publication in the Proceedings of the 2025\\IEEE/RSJ International Conference on Intelligent Robots and Systems (IROS 2025)}}
  \setlength{\headheight}{15pt}
  
}
\newcommand{\vel}{v}

\newcommand{\freS}{\mathfrak{s}}
\newcommand{\freN}{n}
\newcommand{\freU}{u}

\newcommand{\vx}{\vel_x}
\newcommand{\vy}{\vel_y}
\newcommand{\yawRate}{\mathfrak{r}}

\newacronym[plural=MDPs, firstplural=Markov Decision Processes (MDPs)]{mdp}{MDP}{Markov Decision Process}
\newacronym{ekf}{EKF}{Extended Kalman Filter}
\newacronym[plural=UKFs, firstplural=Unscented Kalman Filters (UKFs)]{ukf}{UKF}{Unscented Kalman Filter}
\newacronym[plural=IMUs, firstplural=Inertial Measurement Units (IMUs)]{imu}{IMU}{Inertial Measurement Unit}
\newacronym[plural=RNNs, firstplural=Recursive Neural Networks (RNNs)]{rnn}{RNN}{Recursive Neural Network}
\newacronym[plural=GRUs, firstplural=Gated Recurrent Units (GRUs)]{gru}{GRU}{Gated Recurrent Unit}
\newacronym{lstm}{LSTM}{Long Short-Term Memory}
\newacronym{esc}{ESC}{Electronic Speed Controllers}
\newacronym[plural=GNSSs, firstplural=Global Navigation Satellite Systems (GNSSs)]{gnss}{GNSS}{Global Navigation Satellite System}
\newacronym[plural=OVSs, firstplural=Optical Velocity Sensors (OVSs)]{ovs}{OVS}{Optical Velocity Sensor}

\newacronym{mse}{MSE}{Mean Squared Error}
\newacronym{mae}{MAE}{Mean Absolute Error}
\newacronym{99ae}{99\%-AE}{99th percentile of Absolute Error}

\newacronym{pem}{PEM}{Prediction Error Method}

\newacronym{mlp}{MLP}{Multilayer Perceptrons}

\newacronym{sota}{SOTA}{State-of-the-Art}
\newacronym{rl}{RL}{Reinforcement Learning}
\newacronym{mpc}{MPC}{Model Predictive Control}
\newacronym{ppo}{PPO}{Proximal Policy Optimization}
\newacronym{lerelu}{Leaky ReLu}{Leaky Rectified Linear Unit}

\IEEEoverridecommandlockouts                              

\title{\LARGE \bf
On learning racing policies with reinforcement learning
}

\author{Grzegorz Czechmanowski, Jan Węgrzynowski, Piotr Kicki, Krzysztof Walas
\thanks{
All authors are with IDEAS Research Institute, Warsaw, Poland, IDEAS NCBR, Warsaw, Poland and with Institute of Robotics and Machine Intelligence, Poznan University of Technology, Poznan, Poland. {\tt\small name.surname@put.poznan.pl}
This research was partially funded by PUT internal grant 0214/SBAD/0252. Work of Piotr Kicki was supported by the Foundation for Polish Science (FNP).%
}}

\begin{document}

\maketitle
\thispagestyle{withfooter}
\pagestyle{empty}

\vspace{-0.4cm}

\begin{abstract}

Fully autonomous vehicles promise enhanced safety and efficiency. However, ensuring reliable operation in challenging corner cases requires control algorithms capable of performing at the vehicle limits.
We address this requirement by considering the task of autonomous racing and propose solving it by learning a racing policy using \gls{rl}.
Our approach leverages domain randomization, actuator dynamics modeling, and policy architecture design to enable reliable and safe zero-shot deployment on a real platform. Evaluated on the F1TENTH race car, our \gls{rl} policy not only surpasses a state-of-the-art \gls{mpc}, but, to the best of our knowledge, also represents the first instance of an \gls{rl} policy outperforming expert human drivers in RC racing. This work identifies the key factors driving this performance improvement, providing critical insights for the design of robust \gls{rl}-based control strategies for autonomous vehicles.



\end{abstract}
\vspace{-0.2cm}

\section{Introduction}
Ensuring the reliable performance of fully autonomous vehicles in critical corner cases is essential to achieve their promised safety and enable large-scale deployment. In dynamic scenarios, even small control inaccuracies can lead to crashes. As a result, control strategies must be both robust to uncertainties and capable of operating at the vehicle's limit to minimize the risk of accidents.


Autonomous racing is an example of a task that enables a rigorous comparison of control algorithms operating at the limits of their capabilities.  
The narrow margins of error in racing require vehicles to navigate with exceptional precision, making the racetrack an ideal testbed for evaluating and refining advanced control strategies. To date, classical optimization-based approaches remain among the most widely used methods in autonomous racing~\cite{betz_survey}. Although these methods offer high performance and interpretability, even minor model inaccuracies when operating at the limits of handling can result in crashes. 


\begin{figure}
    \centering
    \includegraphics[width=1.0\linewidth]{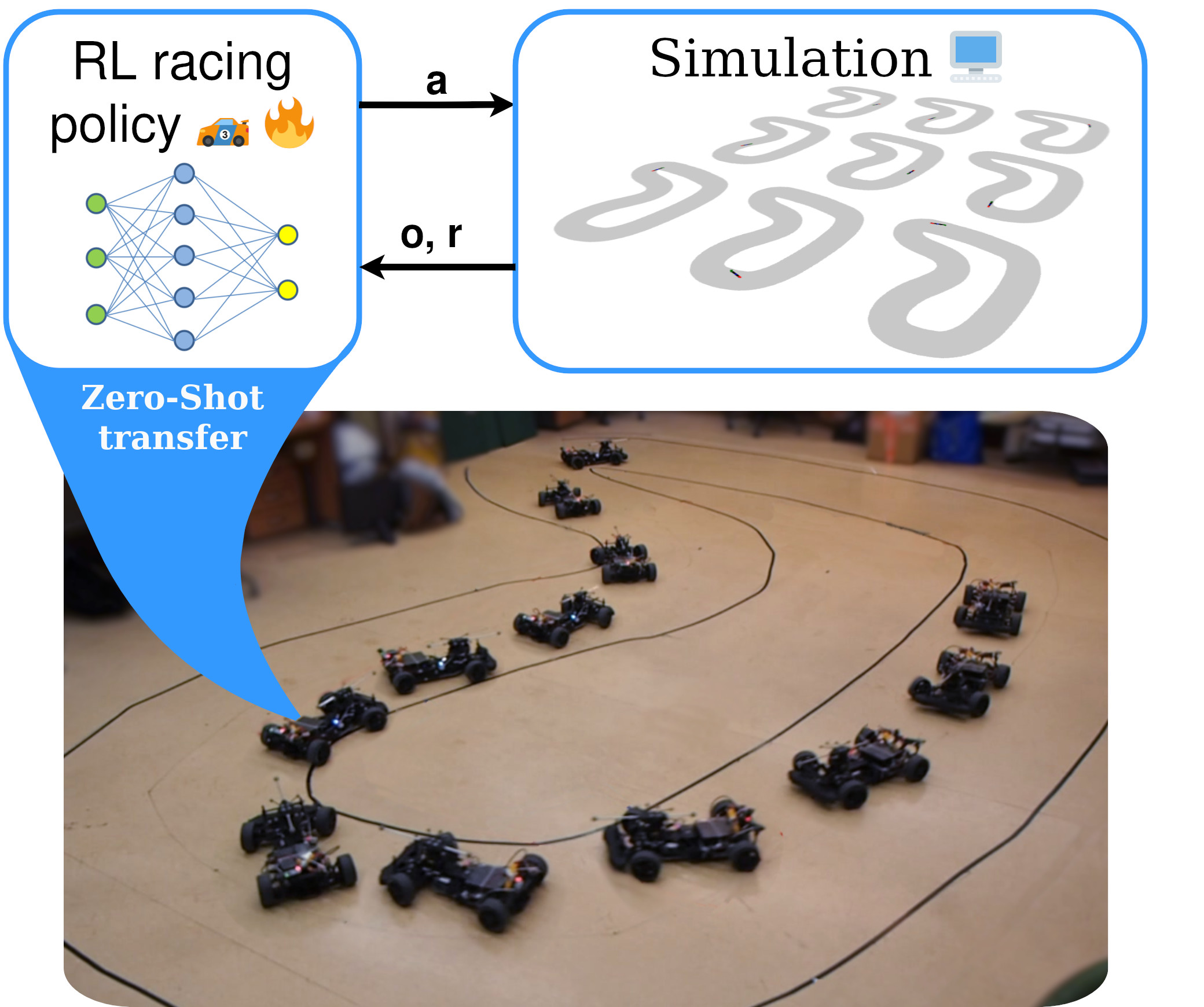}
    \vspace{-0.5cm}
    \caption{Our \gls{rl} policy, trained entirely in simulation, deployed zero-shot on an F1TENTH car in the real world. }
    \vspace{-0.4cm}
    \label{fig:rl_driving}
\end{figure}

One promising approach to developing robust control algorithms is the use of Reinforcement Learning (RL). Unlike optimization-based controllers, \gls{rl} can leverage domain randomization to improve robustness against modeling errors and effectively handle sparse and discontinuous objectives. Moreover, \gls{rl} is not constrained by a finite prediction horizon, allowing it to optimize long-term performance beyond the capabilities of traditional optimization-based methods. Previous studies have demonstrated that \gls{rl} can surpass professional human drivers in car racing simulations~\cite{wurman_outracing_2022},~\cite{song_autonomous_2021},~\cite{fuchs_super-human_2021} and even outperform expert pilots in real-world drone racing~\cite{scaramuzza_champion-level_2023}.
However, despite its success in racing simulations, \gls{rl} policies have struggled to transfer effectively to real-world scenarios. Consequently, optimization-based methods have consistently outperformed \gls{rl} policies in real-world car racing~\cite{ghignone2025rlpp, liniger2021sac}.

In this work, we propose a novel \gls{rl}-based approach to autonomous racing that not only outperforms state-of-the-art \gls{mpc} methods but also, to the best of our knowledge, represents the first instance of a \gls{rl} policy that surpasses an expert human driver in RC racing.  Notably, our benchmark human driver achieved fifth place in the Polish Indoor RC Championship, validating the competitive standard against which our autonomous system was evaluated.
We systematically explore critical design choices, including domain randomization, actuator dynamics modeling, and policy architecture, which are essential to improve performance, safety, and transferability. The \gls{rl} policies are trained exclusively in simulation and subsequently validated on the F1TENTH platform~\cite{f1tenth-v123-o-kelly20a} under real-world conditions without additional training. Supplementary video materials from the experimental evaluation can be found at \href{https://grzegorzczput.github.io/rl-racing/}{https://grzegorzczput.github.io/rl-racing/}.


Our contributions can be summarized as follows:
\begin{enumerate}
    \item We systematically investigated the impact of {domain randomization}, {actuator dynamics modeling}, observation, and action space design on the performance, safety, and transferability of reinforcement learning policies for autonomous racing.
    \item We evaluated our approach against the state-of-the-art \gls{mpc} and an expert human RC driver in the real world F1TENTH racing, demonstrating a fastest lap time improvement of \SI{0.078}{s} over \gls{mpc} and \SI{0.286}{s} over the expert human driver on a challenging \SI{17}{m} racetrack (see Fig.~\ref{fig:rl_driving}).
\end{enumerate}


\section{Related work}
\label{sec:related_work}


\subsection{Model Predictive Control}

One of the most popular approaches to designing a controller for autonomous car racing is to leverage \gls{mpc} algorithms \cite{liniger2015optimization, verschueren2016racingmpc, kabzan_learning-based_2019}. These methods have proven to be effective in real-world deployments. However, a major drawback of optimization-based approaches is their limited control frequency, which is constrained by model complexity and horizon length. This limitation forces users to adopt relatively simple models and significantly shorter planning horizons than a full lap of a racing track.

Additionally, to ensure fast optimization convergence, the objective function typically includes auxiliary regularization terms \cite{kabzan_learning-based_2019, srinivasan_holistic_2021} or requires simplifications, such as restricting costs to quadratic penalties relative to a predefined trajectory \cite{dallas2023adaptiveNon}.
As a result, the objective function often acts as a proxy, rather than directly optimizing for the controller’s true objective.

\subsection{Reinforcement Learning in Autonomous Racing}

On the other hand, \gls{rl} offers a promising solution to overcome these limitations. Reinforcement learning has been successfully applied to complex robotic systems such as quadrupeds \cite{Takahiro2022rslQped, kumar2021rma} and drones \cite{bauersfeld_neurobem_2021, scaramuzza_champion-level_2023}, demonstrating its capability to handle high-dimensional models and long-horizon planning.
Unlike MPC, where the computational burden occurs during inference, \gls{rl} shifts the computational effort to the training phase, enabling efficient real-time execution once training is complete. Moreover, RL is well-suited for optimizing tasks with sparse and delayed rewards, such as minimizing lap times in drone racing \cite{youlong2023RLvsMPC}.

The application of \gls{rl} in autonomous car racing has been studied primarily in simulation \cite{jaritz2018rlsim, ghignone_tc-driver_2023} and has even been shown to outperform professional human simulator drivers \cite{fuchs_super-human_2021}.
Nevertheless, implementing end-to-end \gls{rl} policies in real-world racing remains a non-trivial task. As demonstrated in \cite{ghignone2025rlpp}, researchers could not directly train a high-performance policy without the foundation of a geometric controller. Even with this foundation, the resulting policy still failed to achieve performance comparable to model-based approaches such as \gls{mpc}. Similarly, despite employing real-world policy fine-tuning, the authors \cite{liniger2021sac} were unable to exceed the performance benchmarks set by optimization-based controllers. 



\section{Materials and methods}

\subsection{Problem definition}
In this paper, we consider the problem of real-world autonomous racing with an F1TENTH car. In general, this is an extremely complex problem that typically requires the integration of multiple subsystems responsible for aggregating the data from sensors, performing localization, opponent detection, planning, and control. However, in this paper, we focus solely on the control problem in a time-trial scenario, assuming perfect knowledge of the vehicle state. The goal of this task is to find a controller that, based on the knowledge of the vehicle's state and the racetrack, is able to generate control signals, allowing the car to maximize the progress along the track centerline while remaining within track boundaries. This objective may be considered as an accurate dense approximation of the sparse minimum lap time goal.


\subsection{Proposed solution}

To address the problem of autonomous racing in the time-trial scenario, we begin by developing an accurate simulation of the F1TENTH car. Our goal is to precisely model all relevant aspects of vehicle dynamics, including actuator dynamics and tire characteristics that are lacking in other simulators~\cite{f1tenth_sim-v123-o-kelly20a}. We identify system parameters using a long-horizon prediction loss, which helps stabilize the learned models.


With an accurate simulation in place, we leverage \gls{rl} to train an autonomous racing policy in simulation. Since simulation models are never a perfect representation of reality, deploying the learned policy directly presents challenges due to model mismatches. To enhance sim-to-real transfer, we employ domain randomization, ensuring the policy is robust to simulator inaccuracies and unmodeled behaviors it may encounter in the real world.



\subsection{Hardware setup}
\label{sec:hardware}

\begin{figure}[b]
    \centering
    \includegraphics[width=\linewidth]{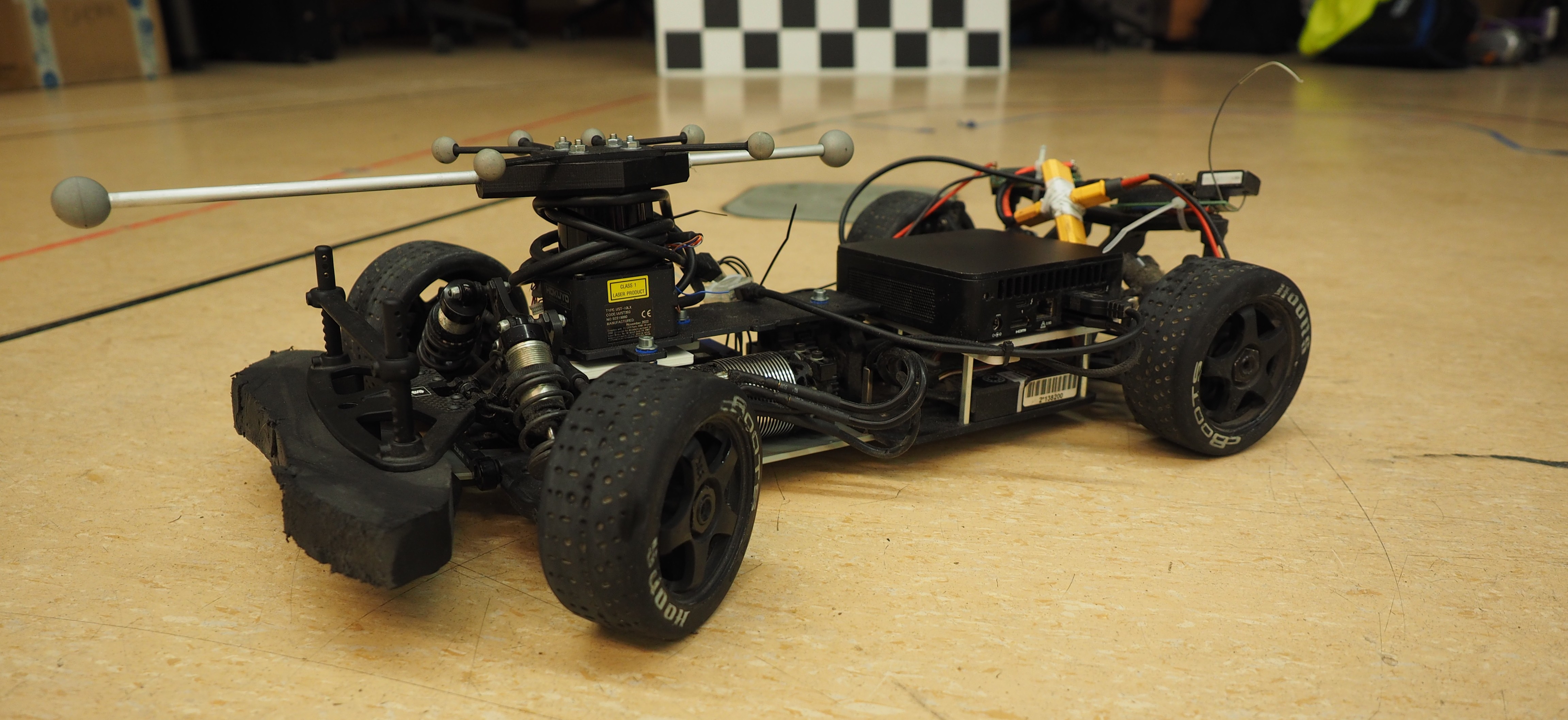}
    \caption{F1TENTH car used for dataset colection and performing real-world experiments.}
    \label{fig:xray}
    
\end{figure}

In this work, we utilized
the F1TENTH platform, built on the Xray GTXE’22—a 1/8th-scale RC car equipped with four-wheel drive and powered by a single motor. Additionally, we equipped the platform with a Dynamixel servo mechanism for the steering column, allowing us to directly measure the steering angle and tune the PID of the position controller. The vehicle operates in two distinct control modes:  

\begin{itemize}
    \item \textbf{Wheel Speed Control Mode}: The control input is defined as $a = [\delta_{\text{ref}}, \omega_{\text{ref}}]^T$, where $\delta_{\text{ref}}$ is the reference steering angle sent directly to the servo for steering actuation, and $\omega_{\text{ref}}$ is the reference wheel speed, which is sent to the \gls{esc}. The \gls{esc} utilizes an internal PID controller to regulate the vehicle's wheel speed accordingly.  
    \item \textbf{Motor Current Control Mode}: The control input is given by $a = [\delta_{\text{ref}}, I]^T$, where the steering control operates identically to the wheel speed control mode. However, instead of controlling wheel speed through the ESC, the motor is directly actuated by setting the current $I$.
\end{itemize}

The two control modes are designed to meet the distinct requirements of human drivers and autonomous control algorithms. Human drivers naturally regulate the vehicle by modulating the throttle signal, which in electric vehicles is typically implemented as motor current control. In contrast, autonomous control systems favor wheel speed regulation using high-frequency, low-level PID controllers inside \gls{esc}.

To enable both system identification and policy deployment, we employ an OptiTrack motion capture system. OptiTrack provides precise real-time measurements of the vehicle’s position and orientation, which are then processed to estimate velocity in the vehicle frame. The velocity is computed by differentiating the position and orientation data using a Savitzky-Golay filter~\cite{savgol}.

All training and real-world experiments were conducted on a single L-shaped track (see Fig.~\ref{fig:rl_driving}), measuring 17 meters in length and 1 meter in width. Despite being constrained by the OptiTrack tracking area, this layout incorporates both a high-speed section and a tight technical sector, providing a balanced environment for evaluating controllers' performance. This challenging setup ensures a comprehensive assessment of reinforcement learning policies and traditional control algorithms.

\subsection{Simulation environment} 
\label{sec/sys_id}
The simulation environment is a key aspect of the reinforcement learning setup, especially for high-performance racing policies. 
In this paper, we focus solely on the accurate and fast simulation of racing car dynamics, completely neglecting the visual aspect of the simulation, as it will only be used for policy training. 
Therefore, we want the dynamics model to be (i) fast to evaluate, (ii) difficult or near impossible to exploit, (iii) accurate enough to enable racing at the limit of grip, and finally (iv) close to the hardware setup we use. 
Such qualities are beyond the scope of general-purpose simulators but can be obtained by using an analytical physics-derived model optimized to be consistent and accurate in long-horizon simulation.

In this paper, we used a single track vehicle dynamics model with the MF6.1 tire model~\cite{besselink2010improved}, which extends the traditional Pacejka formulation~\cite{Pacejka01011992} commonly used in autonomous racing~\cite{kabzan_amz_2019} by jointly considering lateral and longitudinal tire slip to calculate tire forces. Moreover, to account for the dynamics of the physical actuators, we model them as a first-order system, identifying the relationship between the reference input and the measured position and velocity by

\label{sec:methodology}
\begin{equation}    
\begin{aligned}
    \dot{\delta}(t) &= \frac{1}{T_{\delta}} (\delta_{\text{ref}}(t) - \delta(t)), \\ 
    \dot{\omega}(t) &= \frac{1}{T_{\omega}} (\omega_{\text{ref}}(t) - \omega(t)),
\end{aligned}
\label{eq:actuator_dynamics}
\end{equation}
where $T_{\delta}$ and $T_{\omega}$ are the time constants $T_{\delta}$ for the steering actuator and the approximation of the PID regulators inside \gls{esc}, respectively.  

We identified the parameters of the introduced model using the dataset collected with the physical setup introduced in Section~\ref{sec:hardware}. This dataset was collected by an expert human driver operating across various tracks, intentionally varying slip angles and slip ratios to cover the full operational envelope of the vehicle’s state space. 
During data collection, we captured relevant data about the car's internal state, such as motor rotational speed and steering angle, as well as information about its position and orientation using the OptiTrack motion capture system. All of the data were collected synchronously at 100Hz. In total, we recorded 36 minutes of driving.


Finally, similarly to \cite{wegrzynowski2024learning}, we formulated the \gls{mse} loss between the state trajectory roll-out obtained using the dynamics model and ground-truth data on sequences of 85 samples.
To minimize this loss, we used stochastic gradient descent optimization, in particular AdamW~\cite{loshchilov2018decoupled}, and backpropagated the loss gradient with respect to the dynamics model's parameters through time.


\subsection{Reinforcement Learning Training Procedure}  
\label{sec:rl}


Leveraging the accurate simulation environment detailed in Section~\ref{sec/sys_id}, we train our \gls{rl} policy using the \gls{ppo} algorithm~\cite{schulman_proximal_2017, stable-baselines}, in a parallelized setup with 400 simulated instances running concurrently. Each rollout consists of 1024 steps, with a time step of 0.05s. The training was performed using a batch size of 1024. If the agent exceeds track limits during a rollout, it is reset to a random position on the track's centerline to enhance the exploration. During training, the parameters of the dynamics model are randomized at the beginning of each episode by applying multiplicative perturbations with Gaussian noise. Specifically, each parameter $\theta$ is scaled by a factor drawn from $\mathcal{N}(1, \sigma_{dr})$, where $\sigma_{dr}$ is specified in the experimental configuration.

The sum of discounted rewards is calculated with a discount factor of $\gamma = 0.99$. The learning rate decays from $1 \times 10^{-3}$ to $1 \times 10^{-4}$ over the course of training. The total training process spans 120 million environment steps, allowing the agent to experience a wide range of driving scenarios and refine its policy accordingly.

Both the actor and critic networks are implemented as multilayer perceptrons. The actor network consists of two hidden layers, each with 256 neurons, while the critic network comprises two hidden layers with 512 neurons each. Both networks employ the Leaky Rectified Linear Unit activation function with a negative slope of 0.2.

The observation configuration for both networks is defined as follows: 
\begin{enumerate*}
    \item Linear velocities in vehicle frame: longitudinal $v_x$ and lateral $v_y$.
    \item Yaw rate: $\yawRate$.
    \item The angle between the car's heading and the track heading, expressed in Frenet coordinates $\freU$.
    \item Lateral distance from the centerline, in Frenet coordinates: $\freN$.
    \item Measured steering angle $\delta$ and the previous control input $\delta_{\text{ref}}$.
    \item Measured wheel speed $\omega$, commanded wheel speed $\omega_{\text{ref}}$, and the last control input $\dot{\omega}_{\text{ref}}$.
    \item Track information, represented by $N$ points, with curvature $\boldsymbol{c} \in \mathbb{R}^N$ and width $\boldsymbol{w} \in \mathbb{R}^N$. These $N$ points are sampled uniformly in front of the vehicle at 30-centimeter intervals.
\end{enumerate*}  
Thus, the observation vector is given by:
\begin{equation}
    \textbf{obs} = [\vx, \vy,\freU, \freN, \yawRate, \delta, \delta_{\text{ref}}, \dot{\omega}_{\text{ref}}, \omega_{\text{ref}}, \omega, \boldsymbol{c}, \boldsymbol{w}].
\label{eq:obs_space}
\end{equation}
Observations are normalized by dividing each element by its corresponding maximum value. The maximum values for state observations are derived from the system identification dataset, while those for track observations are computed based on the track geometry.

The final output of the actor network is a two-dimensional vector
$[\delta_{\text{ref}}, \dot{\omega}_{\text{ref}}]^T$,
with each element constrained to the interval $[-1,1]$. These outputs are subsequently scaled by a factor of 0.5 for $\delta_{\text{ref}}$ and 5 for $\dot{\omega}_{\text{ref}}$ to yield the action vector. The steering scaling factor was chosen because 0.5 rad is the maximum limit of the steering actuator. The acceleration limit for the wheel was set to 5~$m/s^2$, as this was the highest longitudinal acceleration measured from the identification dataset.
The reference steering angle $\delta_{\text{ref}}$ is directly applied to the servo for steering actuation, while the reference wheel speed derivative $\dot{\omega}_{\text{ref}}$ is integrated over time, analogously to~\cite{djeumou_reference-free_2024}, before being passed to the wheel speed controller. This formulation of speed control mitigates abrupt velocity changes, thereby enhancing policy transferability to the real world.

To effectively train racing policies, we designed the reward function to quantify the vehicle's progress along the track and penalize boundary violations, and formulated it as follows:
\begin{equation}
r_t=\begin{cases}
-1, & \text{if the track boundary is exceeded}, \\
\freS_t-\freS_{t-1}, & \text{otherwise},
\end{cases}
\end{equation}
where $\freS_t$ denotes the progress along the centerline expressed in Frenet coordinates~\cite{frenet}. Although the primary goal is to achieve faster laps, designing the reward function directly around this metric would hinder learning due to its sparse nature. Therefore, progress along the centerline is used as an effective proxy that provides denser feedback. This is why such a reward formulation, or similar variants, is commonly employed in other \gls{rl} racing problems \cite{song_autonomous_2021, fuchs_super-human_2021, scaramuzza_champion-level_2023}.

\subsection{Bridging sim-to-real gap}
An essential aspect of transferring policies trained in simulation to real-world scenarios is to utilize accurate dynamics models.
Nevertheless, even with precise system modeling and identification, discrepancies between simulation and reality are inevitable and can compromise both safety and performance when deploying a policy in real-world conditions. To address this challenge, domain randomization~\cite{domain_rand_open_ai} is often employed during training by perturbing the model parameters. This process enhances the robustness of the learned policy across a range of dynamics models, thereby reducing overfitting to the simulated environment. By training on the distribution of models, the policy becomes less sensitive to specific parameter values and more resilient to unforeseen variations in real-world dynamics, ultimately facilitating a safer and more reliable transfer from simulation to reality.
An important consideration when using domain randomization to enhance robustness is the trade-off between performance and resilience. Although increasing randomization can lead to a more robust policy, it ultimately limits performance because the policy prioritizes conservatism over exploitation, as we show in Section~\ref{sec:domain_rand}.

\section{Experiments}
\label{sec:experiments}

To determine the key factors that affect learning \gls{rl} policies for autonomous racing, we conducted a series of experiments. A major focus of our study was to address the sim-to-real gap, for which we evaluated various domain randomization strategies (see Section~\ref{sec:domain_rand}) and the impact of actuator modeling (see Section~\ref{sec:actuator}) to identify the most effective approach to improve real-world transferability.

Additionally, we investigated the impact of action and observation space choices on policy performance (see Sections~\ref{sec:track} and~\ref{sec:actionspace}), aiming to understand how different representations influence policies' behavior when tested in real-world conditions.

Finally, we compare the introduced RL racing agent with the MPC baseline and refer their results to those obtained by an expert human driver (see Section~\ref{sec:race}).

\subsection{Domain randomization}
\label{sec:domain_rand}

To evaluate the impact of domain randomization on both policy performance and safety, we trained networks using varying levels of parameter perturbation, specifically with $\sigma \in \{0.0,\, 0.02,\, 0.05,\, 0.1\}$. Two randomization strategies were considered: (i) randomizing only the tire-track friction, acknowledging that friction can vary due to small changes in temperature and amount of dust on the track, and (ii) randomizing all single-track parameters (vehicle mass, inertia, weight distribution, etc.) to increase the policy's robustness against model mismatch.

Policy performance was assessed by comparing both lap times and rack boundaries violations, which was quantified by the time-integral of the off-track distance, averaged per lap, measured in $[m \cdot s]$. The off-track distance is defined as
\begin{equation}
e_{\text{off}}(t) = \begin{cases}
|\freN(t)| - W_{\freS(t)}, & \text{if } |\freN(t)| > W_{\freS(t)}, \\
0, & \text{otherwise},
\end{cases}
\end{equation}
where $W_{\freS}$ denotes the track width at the corresponding point $\freS$. The integrated off-track error is given by
\begin{equation}
E_{\text{off}} = \frac{1}{N} \int_{0}^{T} e_{\text{off}}(t) \, dt,
\end{equation}

where $N$ represents the number of laps driven, and $T$ represents the duration of the test episode. During testing, each domain randomization configuration was evaluated over 20 laps (i.e., $N=20$).

\begin{figure}
    \centering
    \includegraphics[width=\linewidth]{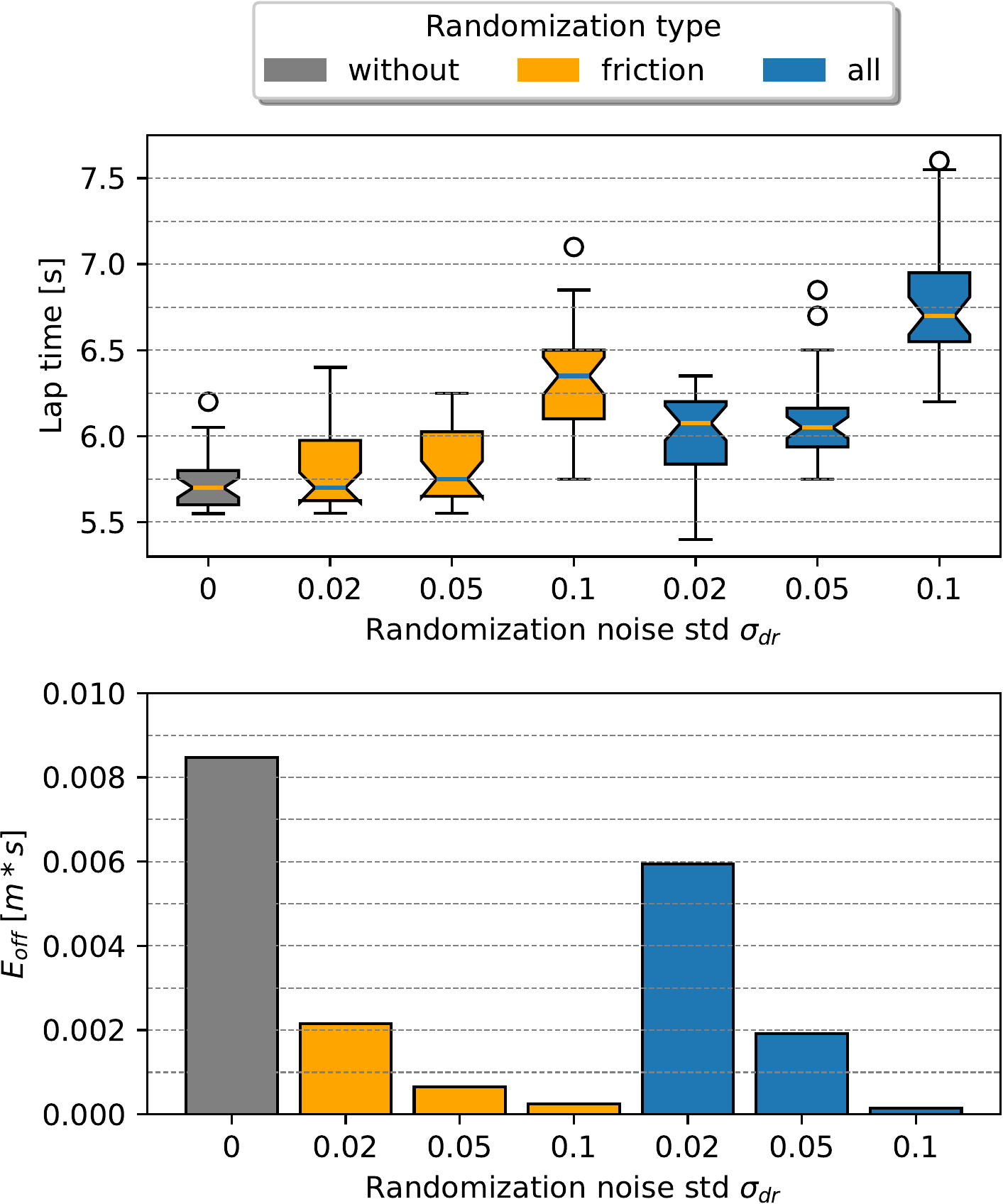}
    \vspace{-0.7cm}
    \caption{Real-world performance of networks trained with randomized friction and single-track parameters. \textbf{Top:} Lap times achieved with different types and $\sigma_{dr}$ of randomization. \textbf{Bottom:} $E_{off}$ and percentage of laps with crashes for different randomization configurations.}
    \vspace{-0.5cm}
    \label{fig:rand_f}
\end{figure}

The results of the experiment with randomized friction levels and single track parameters are presented in Figure~\ref{fig:rand_f}. As friction randomization increases, lap times tend to increase, while the integrated off-track error decreases. A similar effect was observed when randomizing all single-track parameters; however, overall, single-track parameter randomization caused worse performance, resulting in longer lap times and higher $E_{\text{off}}$ compared to policies trained with only friction randomization. This suggests that under more significant parameter variability, the policy adopts a more conservative driving strategy to remain valid across the entire distribution of dynamic models encountered during training. While domain randomization enhances robustness, excessive perturbations across all parameters can degrade overall performance. When the policy is trained over a wide distribution of models, it is forced to learn a control strategy that generalizes across the entire range of dynamics rather than optimizing for a specific one. Since the policy has no access to action history or recurrent memory to infer the underlying system dynamics at a given moment, it cannot adapt its behavior to a particular model instance. Instead, it must adopt a more cautious, suboptimal racing strategy that remains feasible across all possible parameter variations, ultimately leading to longer lap times.

These findings highlight the necessity of balancing domain randomization to ensure both safety and real-world transferability. Even with precise system identification, a moderate level of randomization is beneficial for mitigating simulation-to-reality discrepancies and improving policy robustness. However, while limited domain randomization improves generalization, high-performance policies still require accurate system identification, as randomization alone cannot fully compensate for an inaccurate dynamics model. For subsequent experiments, we selected the policy trained with friction randomization at $\sigma_{dr} = 0.02$, and we refere to it as the base policy. This policy was selected for its acceptable trade-off between speed and robustness.

\subsection{Track representation}
\label{sec:track}

A key aspect of training an effective racing policy is the choice of track representation, which influences learning efficiency and generalization capabilities. Two commonly used approaches to represent track information in \gls{rl} policies are:  

\begin{enumerate}
    \item \textbf{Geometric Representation:} The track is represented using curvature profiles ($\boldsymbol{c}$), track widths ($\boldsymbol{w}$), or range measurements to track boundaries. This provides the policy with information about the upcoming track layout, allowing it to generalize learned behaviors across similar sections. However, this representation has a limited prediction horizon, as it only provides information about the next meters of track, and the policy must simultaneously learn to localize itself on the track from these measurements.
    
    \item \textbf{Track progress representation:} The track information is represented in terms of the Frenet coordinate $s$, which represents progress along the centerline. This approach ensures that the policy always has precise information about its location on the track. However, unlike the geometric representation, it lacks explicit information about the upcoming track shape, limiting the policy’s ability to generalize across similar sections.
\end{enumerate}

To investigate the impact of track representation on policy performance, we trained policies using both approaches: (i) the defaul geometric representation $\textbf{obs}$ defined in \eqref{eq:obs_space} and an alternative (ii) track progress representation $\textbf{obs}_{\freS}$ defined by 
\begin{equation}
    \textbf{obs}_{\freS} = [\freN, \freU, \vx, \vy, \yawRate, \delta, \delta_{\text{ref}}, \dot{\omega}_{\text{ref}}, \omega_{\text{ref}}, \omega, \freS].
\label{eq:obs_space_s}
\end{equation}

Despite the differences in input structure, both policies successfully converged to the same reward levels, as shown in Fig.~\ref{fig:learning}. This suggests that, with sufficient training, \gls{rl} policies can extract the necessary spatial and temporal information from either representation.

Both policies were deployed in real-world experiments, where they achieved comparable performance. However, as presented in Table~\ref{tab:performance_comparison}, the policy utilizing geometric track representation ($\textbf{obs}$) achieved slightly faster lap times.



\subsection{Action space}
\label{sec:actionspace}

\begin{figure}[t]
    \centering
    \includegraphics[width=\linewidth]{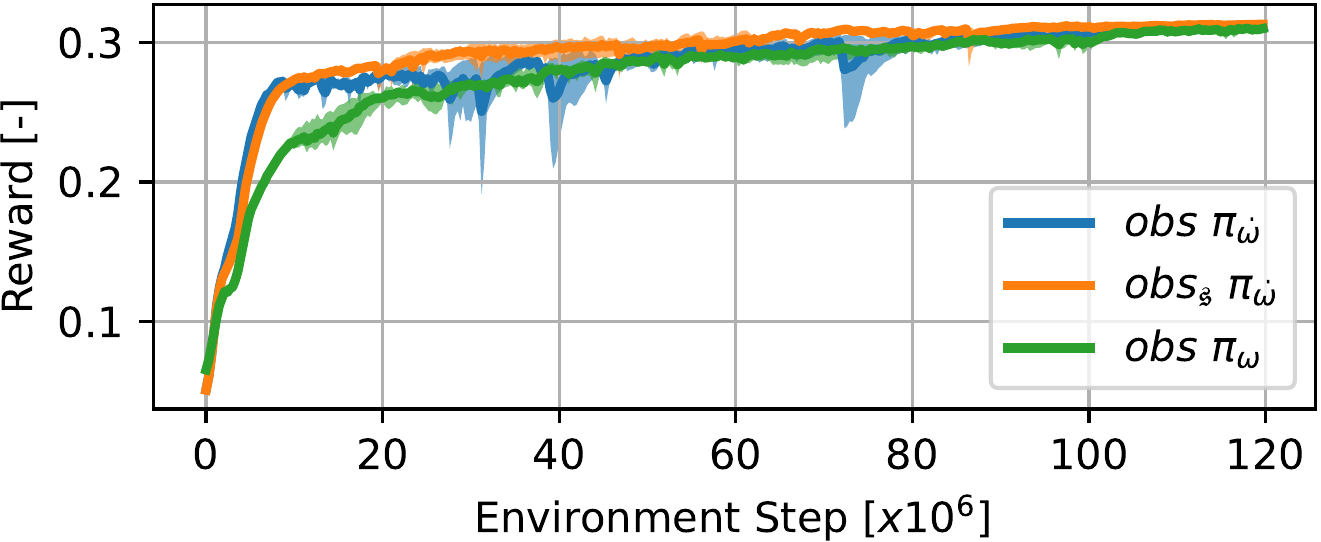}
    \vspace{-0.5cm}
    \caption{Learning curves of the base policy using $\pi_{\dot{\omega}}$ with $obs$, the policy using $\freS$ for track representation $obs_{\freS}$, and the wheel speed control policy $\pi_{\omega}$.}
    \label{fig:learning}
\end{figure}

Another critical policy design choice is the action space selection, which can directly impact both performance and sim-to-real transfer.
We evaluated the effect of control space selection on policy performance and sim-to-real transfer. Using an otherwise identical training configuration, we compared two control architectures: one in which the policy $\pi_{\omega}$ directly controls vehicle wheel speed $\omega_{ref}$ and another in which it modulates the acceleration of the clamped wheel speed $\dot{\omega}_{ref}$. In the simulation, the velocity control policy converged to performance levels comparable to the acceleration control policy, as demonstrated in Figure~\ref{fig:learning}. However, in real-world tests, the wheel speed control-based policy proved overly aggressive, frequently exceeding track boundaries by more than 20 cm and failing to complete 10 laps without crashing. These results underscore the critical importance of choosing a control space that limits abrupt control actions, thus enhancing the robustness of sim-to-real transfer.
\vspace{-0.4cm}

\subsection{Actuator modeling}
\label{sec:actuator}

Actuator dynamics is often neglected in vehicle dynamics modeling. In our work, both the powertrain and the steering actuators are modeled as first-order systems, as described in Equation~\eqref{eq:actuator_dynamics}. To assess the impact of actuator modeling on policy performance, we trained a policy without modeled actuator dynamics, deployed it in the real world, and presented its performance in Table~\ref{tab:performance_comparison}.

\begin{figure}[t]
    \centering
    \includegraphics[width=\linewidth]{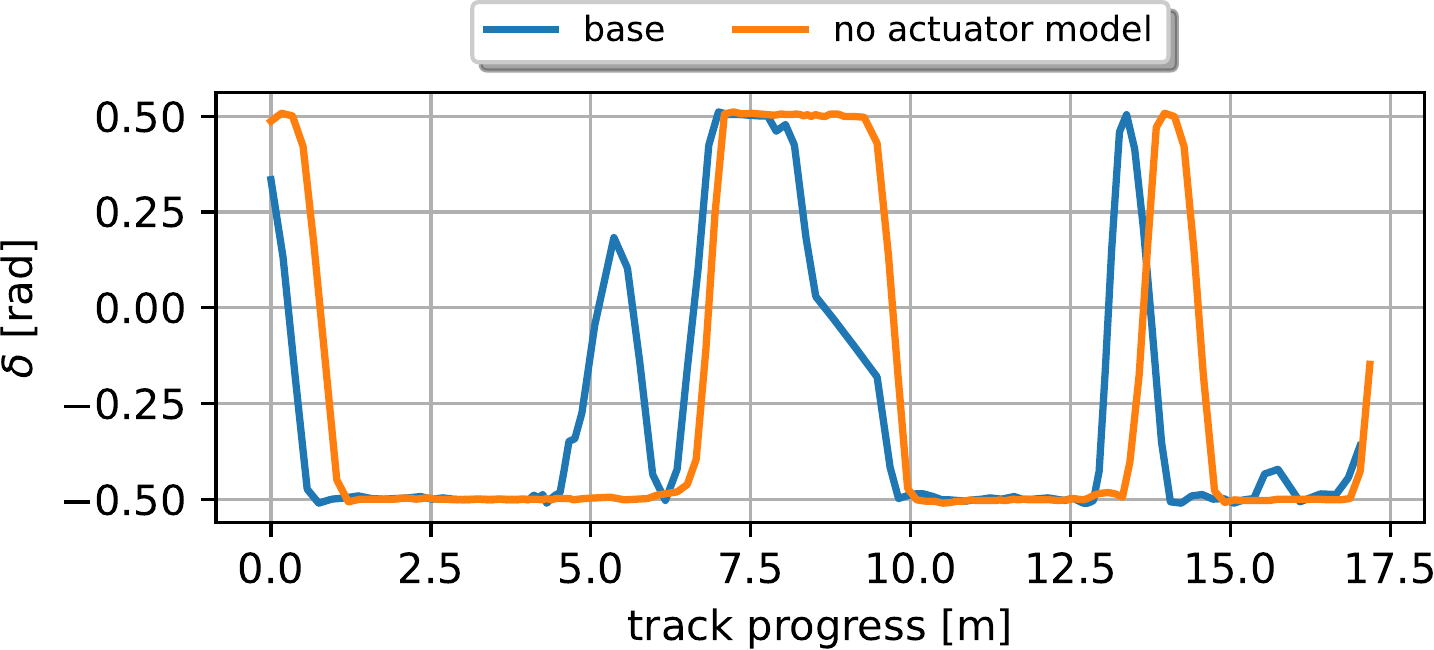}
    \vspace{-0.5cm}
    \caption{Sterring control of networks trained with and without modeled actuators dynamics.}
    \vspace{-0.6cm}
    \label{fig:actuators}
\end{figure}


The policy trained without modeled actuators RL w/o \textit{am} is significantly slower than the baseline RL policy (see Table~\ref{tab:performance_comparison}). As illustrated in Figure~\ref{fig:actuators}, the policy with modeled actuators initiates steering adjustments earlier, effectively compensating for actuator dynamics, while the baseline policy exhibits delayed responses. This finding underscores the importance of incorporating accurate actuator models to improve policy performance and enable more reliable real-world deployment.

\subsection{MPC baseline}

To evaluate our \gls{rl} based controller against other state-of-the-art approaches, we implemented \gls{mpc} using acados~\cite{acados} with a Frenet frame formulation following \cite{verschueren_towards_2014, srinivasan_holistic_2021, novi_real_time_2020}. The controller's objective was to maximize progress along the center line with a soft constraint on track boundaries, similar to~\cite{krinner_mpcc_2024}.
We applied regularization $q_{\beta}(\beta_{kin} - \beta_{dyn})^2$ on the difference between kinematic and dynamic single track model slip angles as in \cite{srinivasan_holistic_2021, novi_real_time_2020}.
To limit the controller's aggressiveness, we incorporated additional action smoothness costs by adding $q_{\delta}(\delta_{\text{ref}} - \delta)^2 + q_{\omega}(\omega_{\text{ref}} - \omega)^2$ to the primary objective. The controller used 80 shooting nodes with a 2.67s horizon, and we tuned the weights $q_{\beta}, q_{\delta}, q_{\omega}$ for optimal performance. To compensate for computational delays, the MPC was initialized from the predicted state.

The MPC controller was implemented with an objective that was as similar to the RL policy as possible, with minimal addition of auxiliary losses to ensure real-time optimization convergence. Furthermore, the dynamics model used in the MPC was identical to the one in the simulation environment, resulting in our best-performing MPC implementation.

\subsection{Policy performance}
\label{sec:race}

\begin{figure*}[t]
    \centering
    \includegraphics[width=\textwidth]{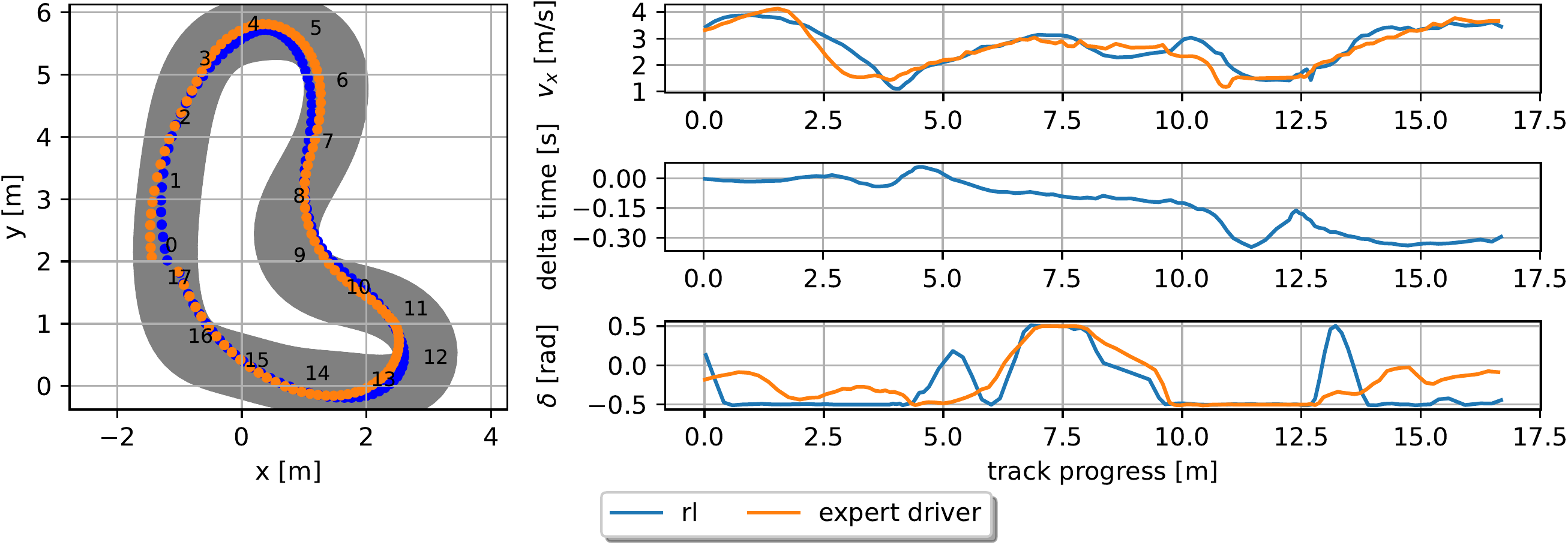}
    \caption{Comparison of the fastest recorded lap without exceeding track boundaries by a human expert driver and the \gls{rl} policy. \textbf{Left:} Trajectories driven on the track, with black numbers indicating progress along the center line $\freS$. \textbf{Right:} From top to bottom, the panels display the car’s longitudinal velocity ($v_x$) during the lap, the delta time between the \gls{rl} policy and the expert human driver, and the variation in steering command.}
    \label{fig:rl_vs_human_img}
    \vspace{-0.2cm}
    
\end{figure*}

To evaluate our racing policy, we compared its performance against a state-of-the-art \gls{mpc} approach and an expert human driver, who placed fifth in the Polish Indoor RC Championships. 

To ensure a fair comparison between the autonomous approaches and the human driver, we provided the expert with the track layout a week before the test. The driver was allowed to test the car in both the Motor Current Control Mode and Wheel Speed Control Mode. After evaluating both options, he selected Current Control Mode and was allowed to adjust the maximum current limit to match his driving style. 

Following this setup, he was given two hours of practice to familiarize himself with both the car and the track. He controlled the car from an elevated position above the track surface to ensure optimal visibility.

\begin{table}[t]
    \centering
    \caption{Comparison of Fastest Lap Times, Average Lap Times, and ${E}_{\text{off}}$. Over a 20-Lap Experiment.}
    \label{tab:performance_comparison}
    \renewcommand{\arraystretch}{1.1} 
    \setlength{\tabcolsep}{6pt}
    \begin{tabular}{lccc}
        \toprule
        \textbf{Name} & \textbf{Fastest Lap* (s)} & \textbf{Avg Lap Time (s)} & \textbf{$\boldsymbol{E}_{\text{off}}$ (m$*$s)}  \\
        \midrule
        RL $\textbf{obs}$                       & \textbf{5.561} & \textbf{5.768}$\pm$\textbf{0.10}  & 0.00216  \\
        RL w/o  \textit{am}**       & 6.430  & 6.785$\pm$0.22  & 0.00118  \\
        RL $\textbf{obs}_{\freS}$                 & 5.598  & 5.812$\pm$0.12  & 0.00178  \\
        MPC                             & 5.639  & 5.804$\pm$0.17  & \textbf{0.00083}  \\
        Human                           & 5.847  & 5.939$\pm$0.19  & 0.01970  \\
        \bottomrule
    \end{tabular}
    \vspace{0.4cm}
    \footnotesize{\\ * fastest lap time recorded without exceeding track boundaries. \\
                    ** RL policy trained without actuator modeling}.
    \vspace{-0.4cm}

\end{table}


After the practice session, the driver was instructed to complete 20 laps as quickly as possible while staying within track boundaries. The performance metrics of the human driver, our \gls{rl} policy, and the \gls{mpc} approach are summarized in Table~\ref{tab:performance_comparison}.


The \gls{rl} policy outperformed the \gls{mpc} baseline in both fastest-lap and average-lap times. We attribute \gls{rl}’s advantage, at least in part, to \gls{mpc}’s finite prediction horizon, which biases control toward locally optimal actions rather than lap-time–optimal trajectories. Moreover, the \gls{rl} policy surpassed the expert human driver across all evaluated metrics.To the best of our knowledge, this is the first time that the \gls{rl} policy has exceeded expert human performance in RC car racing.
Notably, \gls{mpc} achieved the lowest $\boldsymbol{E}_{\text{off}}$, while the human driver recorded the highest. This discrepancy may be attributed to the track's relatively small size compared to the vehicle and the low tire-road friction conditions, which differed significantly from those typically encountered in RC racing competitions. Unlike traditional RC race tracks, which are often carpeted to enhance grip, our test track's slippery surface may have posed an unexpected challenge for the human driver, limiting their ability to fully exploit the vehicle’s handling capabilities.

 The racing lines followed by the human driver and the \gls{rl} policy — illustrated in the left image of Figure~\ref{fig:rl_vs_human_img} — are similar. Although the \gls{rl} policy initially takes a tighter arc from $\freS = 0~m$ to $\freS =5~m$, it does not immediately gain time over the human driver, but instead positions the vehicle more favorably for the upcoming sections. Between $\freS = 5~m$ and $\freS =11~m$, while both maintain the same longitudinal velocity ($v_x$), the \gls{rl} policy gains approximately 0.14 seconds by adopting a more efficient racing line. In the final segment, from $\freS = 11~m$ to $\freS =17~m $, the policy carries more speed into the corner, securing an additional 0.15-second advantage, resulting in a total lead of nearly 0.3~seconds over the expert human driver for the entire lap.

This comparison reinforces the notion that \gls{rl}-based methods can surpass not only expert human drivers but also state-of-the-art optimization-based controllers.
These results highlight the potential of reinforcement learning as a viable alternative for high-performance autonomous racing and, more broadly, for real-world applications that require control at the vehicle limits.



\section{Conclusion}
\label{sec:conclusion}

In this work, we investigated the applicability of reinforcement learning to real-world autonomous racing using a scaled car, evaluating its ability to operate at the limits of vehicle dynamics while maintaining safety and transferability to real-world deployment. Our approach systematically explored the impact of domain randomization, actuator modeling, and policy architecture on both performance and robustness. To the best of our knowledge, we demonstrated that our \gls{rl} policy is the first to surpass state-of-the-art \gls{mpc} and expert human driver in RC car racing.

Our findings highlight several key insights for \gls{rl}-based racing policies. First, moderate domain randomization improves robustness and sim-to-real transferability; however, excessive perturbations across all parameters degrade performance by forcing the policy to generalize too broadly. Additionally, accurate actuator modeling is critical for achieving high performance, as it enables the policy to anticipate actuator dynamics and optimize control actions accordingly. Finally, selecting an appropriate control architecture plays a crucial role in stability, with action spaces that limit abrupt control changes, leading to safer and more effective real-world policies.

Despite these advancements, several challenges remain. Although our policy demonstrates superior performance in single-lap racing, future work should focus on adapting to dynamically changing track conditions and incorporating real-time model adaptation. Moreover, further exploration of hybrid learning-based and model-based approaches may yield even greater improvements in performance and safety.

Our results underscore the potential of \gls{rl} for high-performance autonomous racing and provide valuable guidelines for future research in safe and efficient learning-based control strategies.

\newpage






\bibliographystyle{IEEEtran}
\bibliography{manual_references}

\end{document}